# The best way to select features?
## Comparing MDA, LIME and SHAP

Xin Man
QTS Capital Management, LLC.
Niagara-on-the-Lake, ON, Canada
manxin0821@gmail.com

Ernest P. Chan
QTS Capital Management, LLC.
Niagara-on-the-Lake, ON, Canada
ernest@qtscm.com

**ABSTRACT**

Feature selection in machine learning is subject to the intrinsic randomness of the feature selection algorithms (e.g. random permutations during MDA). Stability of selected features with respect to such randomness is essential to the human interpretability of a machine learning algorithm. We proposes a rank-based stability metric called 'instability index' to compare the stabilities of three feature selection algorithms MDA, LIME, and SHAP as applied to random forests. Typically, features are selected by averaging many random iterations of a selection algorithm. Though we find that the variability of the selected features does decrease as the number of iterations increases, it does not go to zero, and the features selected by the three algorithms do not necessarily converge to the same set. We find LIME and SHAP to be more stable than MDA, and LIME is at least as stable as SHAP for the top ranked features. Hence overall LIME is best suited for human interpretability. However, the selected set of features from all three algorithms significantly improves various predictive metrics out-of-sample, and their predictive performances do not differ significantly. Experiments were conducted on synthetic datasets, two public benchmark datasets, and on proprietary data from an active investment strategy.

## 1 Introduction

Currently, many feature selection algorithms in machine learning suffer from the 'random seed' problem. If we perform feature selection on a prediction model multiple times with different seeds, a feature in a run may be ranked as the 'most important feature' but dropped to a low rank in another run. This is problematic because many researchers rely on manual inspection of the top selected features from a machine learning algorithm to build intuition and trust of the algorithm, and in fact the selected features are sometimes the only desired output of a machine learning program. If every random seed produces a different set of selected features, the output is not interpretable.

There are many existing measures of stability discussed by [1-3]. Our proposed stability measurement calculates the stability of each feature separately, and overall stability is the *rms* of the feature stability scores across all features. The stability score is derived from the variance of a feature's ranks across iterations and the higher the score the lower the stability. Hence we call it the 'instability index'.

Feature importance score indicates how much information a feature contributes when building a supervised learning model. The importance score is calculated for each feature in the dataset, allowing the features to be ranked. The MDA importance score [4] is measured by the Mean Decrease Accuracy of a random forest when the values of a feature are permuted in the out-of-bag samples. Another method called LIME [5] locally explains 'Black-Box' classifiers with a linear regression model. The absolute value of the regression coefficient of a feature is taken as the importance score of that feature for a sample. Using ideas from coalitional game theory, the SHAP method [6] computes the Shapley value of a feature, which is the average of the marginal contributions of that feature to all predictions across all permutations of selected features.

Splitting the dataset into train, validation, and test sets, a random forest model is trained on the train set. With this trained model, feature selection is performed on the validation set. Using only the selected features, a new random forest is trained and its out-of-sample performance is measured on the test set. We compare the three algorithms using F1 score, AUC and Accuracy for classification problems and using MSE, MAE and R2 for regression problems. For our trading strategy, we also compare financial metrics including Sharpe ratio and returns.

Applying these methods to the two synthetic datasets, two public datasets, and our proprietary financial trading dataset, we will see that
1. LIME and SHAP are consistently more stable than MDA;
2. LIME is at least as stable as SHAP for the top ranked features, sometimes more, hence better for human interpretation;
3. The number of iterations used in feature selection needs to be large enough to minimize variability;
4. The selected subset of features achieves better predictive performance compared to using the full feature set, and all three algorithms achieve similar predictive performance;
5. The selected subset also improves the financial metrics of our trading strategy.

The rest of this paper is organized as follows: Section 2 defines 'instability index' which is used to evaluate the stability of the feature importance scoring algorithms MDA, LIME, and SHAP ; Section 3 compares instability of these algorithms in two synthetic and two public datasets; Section 4 discusses if the predictive performance can be improved by feature selection; Section 5 investigates the convergence property of these algorithms and the relation between convergence and predictive performance; Section 6 applies our findings to the improvement of our trading strategy.

## 2 Instability Index

Most feature selection algorithms involve randomness: if we start from different random seeds, it is *not* guaranteed that the importance score or rank of each feature remains the same and that the same features are selected each time. This randomness is due to the random permutations of the values of one feature at a time for MDA, the random perturbations of all the features at the same time for each sample for LIME, and the random permutations of the feature sequence in both forward and reverse directions for SHAP. We define an instability index to evaluate how randomness affects the rankings of the important features in each algorithm.

Applying a feature scoring algorithm, we can obtain an n×m importance scores matrix S from an m-feature dataset with n iterations

$$S = \begin{bmatrix} s_{11} & \cdots & s_{1m} \\ \vdots & \ddots & \vdots \\ s_{n1} & \cdots & s_{nm} \end{bmatrix},$$

where $s_{ij}$ is the importance score for the $j^{th}$ feature for the $i^{th}$ iteration. Each iteration is based on a different random permutation of the rows of a feature in MDA, a random perturbation of the features of a sample in LIME, or a random permutation on the order of features in SHAP.

Denote R as the corresponding rank matrix of S given by

$$R = \begin{bmatrix} r_{11} & \cdots & r_{1m} \\ \vdots & \ddots & \vdots \\ r_{n1} & \cdots & r_{nm} \end{bmatrix},$$

The ranks of features of a run are obtained by sorting their importance scores in that run and assigning a rank of 1 to the highest score (the most important feature). The average rank of the n iterations is

$$r_j = \frac{r_{1j} + \cdots + r_{nj}}{n},$$

The feature importance of feature j is measured by the reciprocal of $r_j$ and then normalized so that the sum is 1 for all features:

$$\tilde{r}_j = \frac{\frac{1}{r_j}}{\frac{1}{r_1} + \cdots + \frac{1}{r_m}} \tag{1}$$

Note this rank-based feature importance score is independent of the specific feature selection algorithm, such as MDA, LIME, or SHAP. This makes it easy to define the stability (or instability) of a feature for any selection algorithm.

The 'instability' of the feature j is defined as its variance

$$V_j = Var(r_{1j}, \dots, r_{nj}).$$

The 'instability index' of a feature selection algorithm for a dataset is calculated from the average of the top k features' instability scores. Hence, the 'instability index' is

$$I = \sqrt{\frac{V_{(1)} + \cdots + V_{(k)}}{k}}, \tag{2}$$

where $V_{(k)}$ is the variance of the $k^{th}$-most important feature. We will study how the instability index changes with k.

In this paper, our SHAP implementation is derived from Lundberg's PermutationExplainer[1] which is based on the KernelExplainer. If a model contains M features, there will be $2^M$ possible coalitions. To economize, KernelExplainer samples a smaller subset when M is large. But SHAP also has the TreeExplainer specifically for tree-based models that provides a deterministic result for feature rankings. Hence the instability index of TreeExplainer is always zero for a fixed random forest. But if we apply cross validation during feature selection, the random forest will differ for each validation fold, giving rise to different rankings of features across various validation folds even if we used TreeExplainer. Hence to better simulate the effect of random seeds on SHAP, and to make the comparisons more relevant to machine learning algorithms besides random forests, we choose PermutationExplainer for our study.

## 3 Instability Comparisons

To compare the stability among the feature scoring algorithms, we construct the data matrix (X,y) from two synthetic datasets, two public datasets, and a proprietary data set derived from our trading strategy's performance. The label y is either a binary or a continuous variable, and random forest classification or regression is used accordingly. If other machine learning models are used, our conclusions may well change.

The train/validation/test split is 0.6/0.2/0.2. The train set is used to train the random forest with the entire feature set, the validation set is used for feature selection, and the prediction performance is evaluated on the test set with selected features.

### 3.1 Synthetic Data

The synthetic dataset can be generated from the 'Scikit-learn' module in Python. To test how the proposed method responds to synthetic data, the dataset is composed of three kinds of features. They are described in [7] as: 1. Informative features that are used to determine the label; 2. Noisy features that bear no information on determining the labels; 3. Redundant features which are linear combinations of the informative features. Each synthetic dataset has 1000 samples and 40 features including 10 informative, 10 redundant and 20 noisy features. These features are named 'I_*' as informative, 'N_*' as noise, and 'R_*' for redundant.

---
[1] https://github.com/slundberg/shap

### 3.1.1 Classification

The synthetic data has two classes with sample sizes 503 and 497 which is very close to a 'balanced dataset'. We compare MDA, LIME, and SHAP by including the top k features for k= 1, …, 40. (We do not worry about optimizing k yet.) As shown in Figure 1, the instability index of MDA computed on the validation set is consistently higher than that of LIME and SHAP, which means MDA is the least stable method. LIME is more stable than SHAP for top-ranked features but SHAP is more stable when most features are included. Note that even when all features are selected by all three algorithms, they still differ in their instability index because they rank the features differently. But of course, the test set predictive performances in Figure 3 will then be identical across the three algorithms.

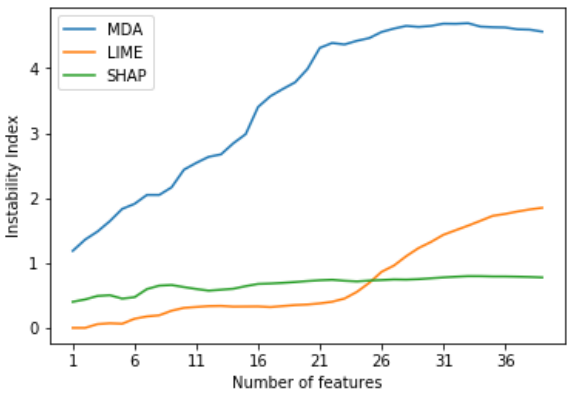

Figure 1: Instability index comparison for synthetic classification dataset

We can take a closer look at the top-ranked feature. From Figure 2, 'R_4' is ranked 1st in all 100 iterations using LIME while 'R_7' is placed top in about 80 iterations using SHAP. Feature 'I_0' selected by MDA has a flatter distribution. It corroborates the conclusion that LIME is more stable for top-ranked features in this dataset. It is also sensible that a redundant feature is picked, as it incorporates information from multiple informative features.

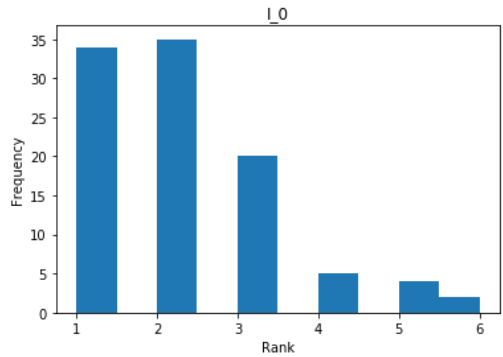

(a) MDA

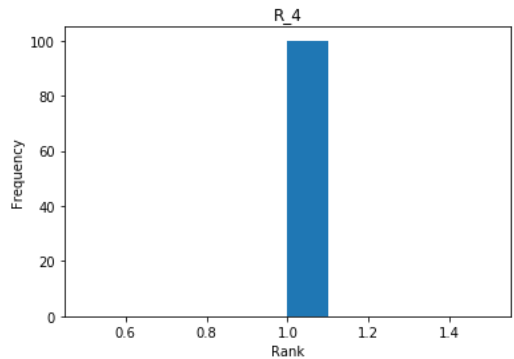

(b) LIME

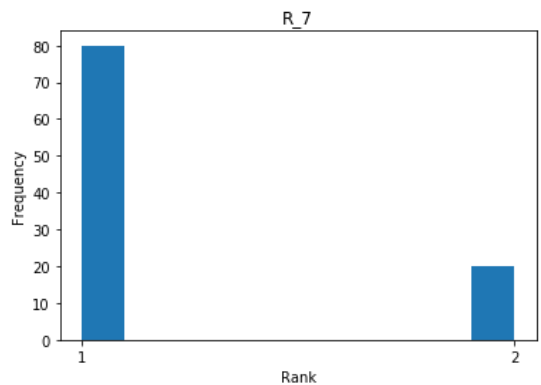

(c) SHAP

Figure 2: Histogram of the highest scored feature for synthetic classification dataset

Although LIME may generate the most stable features ranking, it does not necessarily mean that it outperforms others in terms of predictive performance metrics. From Figure 3, these algorithms perform similarly in AUC in the test set. (The plots of F1 and Acc are similar and thus omitted.) As all three AUC curves peak at some intermediate number of features, selecting an optimal subset of features can bring a better model performance than including the entire feature set. We will see in Section 4 that this is true for most datasets.

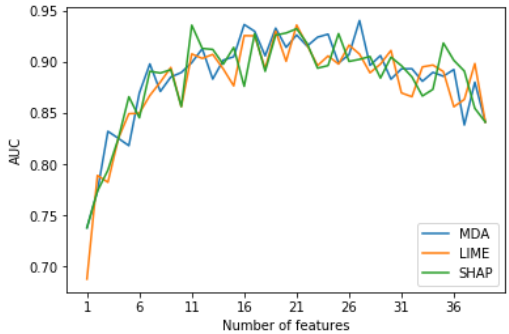

Figure 3: Prediction performance for synthetic classification dataset on test set

### 3.1.2 Regression

When the label y is a continuous variable, we apply random forest regression. Similar to the classification example, Figure 4 shows MDA is consistently the least stable method, and SHAP is the most stable when many features are included.

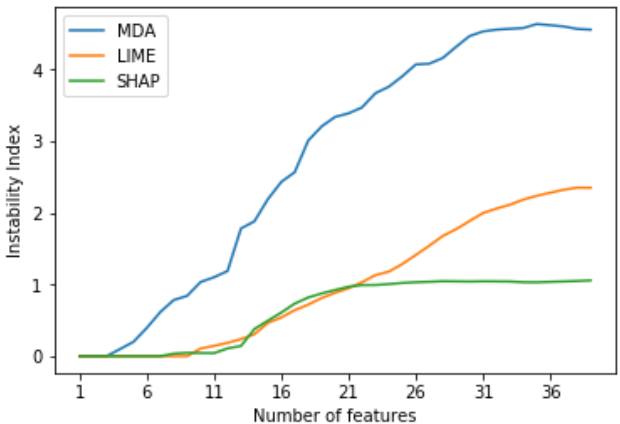

Figure 4: Instability index comparison for synthetic regression dataset

In this dataset, every feature importance method ranks the feature 'R_3' at the top in all the 100 iterations. Note that once again a redundant feature is picked.

To evaluate prediction performance on the test set, the criteria chosen were mean absolute error (MAE), mean squared error (MSE), and R-squared. As with the classification example, all performance curves reach their best values at some intermediate number of features and therefore selecting an optimal subset of features can bring better model performance than including the entire feature set. We display the MSE in Figure 5 as an example.

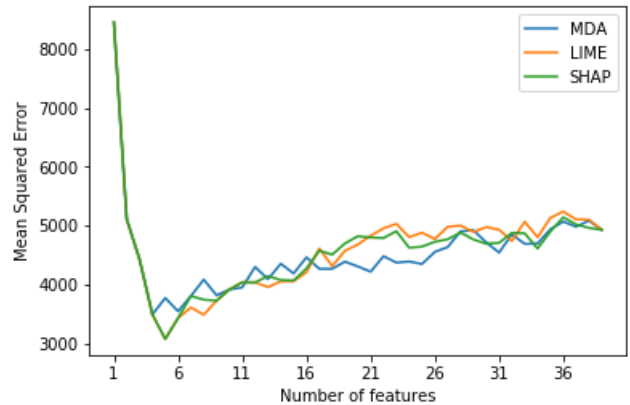

Figure 5: Prediction performance for synthetic regression dataset on test set

## 3.2 Public Data

Besides simulated datasets, two public datasets are also analyzed, one for classification and one for regression.

### 3.2.1 Breast Cancer Dataset

The breast cancer dataset[2] is a binary classification dataset with sample size 569 and feature size 30. The features are computed from a digitized image of a fine needle aspirate of a breast mass which describe characteristics of the cell nuclei present in the image. The target variable is if the cancer is malignant or benign. Once again MDA is consistently the least stable method, but LIME and SHAP have very similar stability for all choices of number of features as shown in Figure 6.

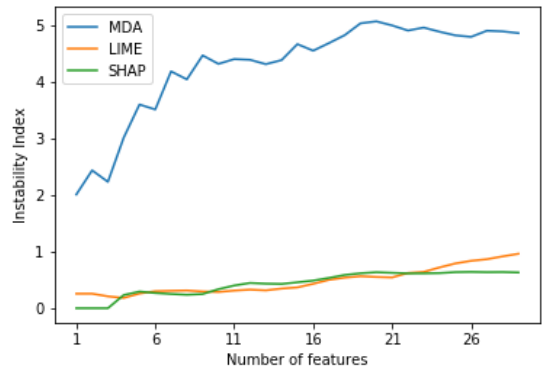

Figure 6: Instability index comparison for Breast Cancer dataset

The feature 'worst radius' is ranked as the most important feature by MDA while 'worst concave points' is selected by both LIME and SHAP. While the rank of 'worst radius' varies widely across different MDA iterations, 'worst concave points' ranks first for all 100 SHAP iterations and more than 90 LIME iterations.

---

[2] https://archive.ics.uci.edu/ml/datasets/Breast+Cancer+Wisconsin+(Diagnostic).

The prediction performances of the three algorithms are evaluated by F1 score, AUC and accuracy, and they are very similar. More details of the results can be seen in Section 4.

### 3.2.2 Boston Housing Price

The Boston Housing dataset[3] contains 506 samples and 13 features. The features are factors related to the housing market and the target variable is the median value of a home. Once again MDA is consistently the least stable method, and LIME is more stable than SHAP when the number of features is small, as shown in Figure 7.

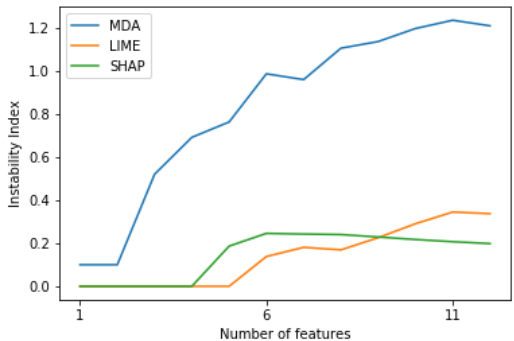

Figure 7: Instability index comparison for Boston Housing Price dataset

The feature 'LSTAT' occupies first place for all three algorithms. This feature appears to be very stable as it is ranked in the first place for all iterations by LIME and SHAP and for the majority of iterations by MDA.

Again, the prediction performances of the three feature selection algorithms are very similar.

## 4 Does feature selection improve predictive performance?

As suggested in [7], we select the top k ranked features with importance scores higher than the mean importance scores across all features. Using the selected features in random forest models, we retrain a random forest, and its prediction performance on the out-of-sample test set is summarized in Table 1 below.

Table 1: Summary Table for Prediction Performance on Test Set

|  | Synthetic Classification | | |
|---|---|---|---|
|  | F1 | AUC | Acc |
| MDA | 0.791 | 0.899 | 0.805 |
| LIME | 0.814 | 0.856 | 0.820 |
| SHAP | 0.814 | 0.856 | 0.820 |
| All | 0.778 | 0.841 | 0.795 |
|  | Synthetic Regression | | |
|  | MAE | MSE | $R^2$ |
| MDA | 49.56 | 3812.07 | 0.895 |
| LIME | 49.82 | 3911.96 | 0.963 |
| SHAP | 49.82 | 3911.96 | 0.892 |
| All | 57.14 | 4926.75 | 0.864 |
|  | Breast Cancer (Classification) | | |
|  | F1 | AUC | Acc |
| MDA | 0.981 | 0.982 | 0.974 |
| LIME | 0.987 | 0.988 | 0.982 |
| SHAP | 0.961 | 0.980 | 0.947 |
| All | 0.954 | 0.980 | 0.939 |
|  | Boston Housing (Regression) | | |
|  | MAE | MSE | $R^2$ |
| MDA | 2.56 | 13.15 | 0.822 |
| LIME | 2.59 | 13.68 | 0.815 |
| SHAP | 2.52 | 13.28 | 0.820 |
| All | 2.54 | 14.00 | 0.811 |

Table 1 shows that for all data sets except the Boston Housing data, all feature selection algorithms outperform predictions using all features ("All" in our table). As Boston Housing data contains only 13 features, feature selection may not improve the prediction when the full feature set is already small. The predictive performance differences among the various selection algorithms are minor, despite significant differences in their instabilities.

## 5 Convergence

In the implementation of MDA, every feature is permuted multiple times and, following the Python Scikit-learn library, we call the number of permutations 'n_repeat' (with a default of 5). In LIME, each instance and its perturbed samples only fit one linear model, so effectively the default 'n_repeat' is 1 (though this hyperparameter isn't defined in LIME's standard implemenation). The argument 'n_permutation' in SHAP's PermutationExplainer represents the number of permutations with a default value of 1. This is also effectively our n_repeat.

Do the selected features converge to a fixed set when the number of iterations 'n_repeat' is large? To investigate this, we run 10 experiments on the four datasets discussed in Section 3, and in each experiment n_repeat ranged from 1 to 1000. Thus for any n_repeat, each feature gets 10 rankings after the experiments. We expect to see the variance decreases to zero as n_repeat increases to infinity. To represent the 'variance' of 10 sets of feature ranks, we compute the instability index.

Figure 8 shows the instability index from 1 to 1000 iterations. Except for the Boston Housing dataset, the instability index for the other three datasets (exemplified by the Synthetic Classification Data's curve displayed in the figure) ends up with a nonzero value which means the selected features do not converge to a unique set. This lack of convergence also cannot be explained by the substitution effect as [8] alone, as we tried removing the redundant features in the synthetic data and the non-uniqueness persists.

---
[3] https://www.cs.toronto.edu/~delve/data/boston/bostonDetail.html.

Although the instability index does not converge to zero in three out of four datasets, it monotonically decreases. From Figure 8, the convergence speed of SHAP and LIME on the Synthetic Classification data is faster than MDA and their instability index is also consistently lower than MDA, which implies the 10 feature sets selected by LIME or SHAP are less different from each other than those selected by MDA. The same behavior holds true for all other data sets except for the Boston Housing data – the latter is separately plotted in Figure 8.

We find that the absolute value of the slope in Figure 8 as well as the two other omitted datasets is smaller than 0.01 at 'n_repeat ≤ 100'. Hence, 'n_repeat = 100' was used for all the experiments in this paper.

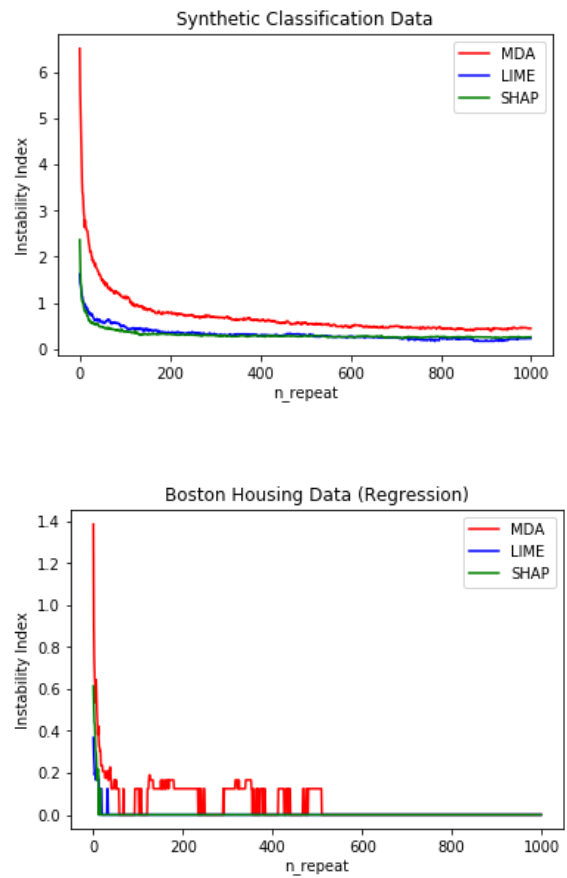

Figure 8: Instability index versus 'n_repeat'

Do more iterations also improve prediction performance? In Table 2, we compare the prediction results obtained with different number of iterations. Except for the Boston Housing Price regression dataset, there is no evidence that a larger number of iterations can make a better prediction. Increasing the number of iterations may only increase feature stability rather than improve the predictive performance.

Table 2: Prediction Performance Comparison for Various Iterations on Test Set

(a) Synthetic Classification

|  | Default | | |
|---|---|---|---|
|  | F1 | AUC | Acc |
| MDA | 0.798 | 0.891 | 0.800 |
| LIME | 0.749 | 0.836 | 0.755 |
| SHAP | 0.800 | 0.908 | 0.805 |
|  | 100 Iterations | | |
|  | F1 | AUC | Acc |
| MDA | 0.791 | 0.899 | 0.805 |
| LIME | 0.814 | 0.856 | 0.820 |
| SHAP | 0.814 | 0.856 | 0.820 |
|  | 1000 Iterations | | |
|  | F1 | AUC | Acc |
| MDA | 0.827 | 0.922 | 0.830 |
| LIME | 0.754 | 0.846 | 0.765 |
| SHAP | 0.833 | 0.901 | 0.840 |

(b) Synthetic Regression

|  | Default | | |
|---|---|---|---|
|  | MAE | MSE | $R^2$ |
| MDA | 50.57 | 4282.58 | 0.901 |
| LIME | 51.12 | 4485.15 | 0.896 |
| SHAP | 50.46 | 4272.79 | 0.901 |
|  | 100 Iterations | | |
|  | MAE | MSE | $R^2$ |
| MDA | 49.56 | 3812.07 | 0.895 |
| LIME | 49.82 | 3911.96 | 0.963 |
| SHAP | 49.82 | 3911.96 | 0.892 |
|  | 1000 Iterations | | |
|  | MAE | MSE | $R^2$ |
| MDA | 48.63 | 3793.32 | 0.891 |
| LIME | 42.74 | 2901.66 | 0.916 |
| SHAP | 49.81 | 4010.32 | 0.884 |

(c) Breast Cancer (Classification)

|  | Default | | |
|---|---|---|---|
|  | F1 | AUC | Acc |
| MDA | 0.951 | 0.972 | 0.930 |
| LIME | 0.970 | 0.964 | 0.956 |
| SHAP | 0.970 | 0.964 | 0.956 |
|  | 100 Iterations | | |
|  | F1 | AUC | Acc |
| MDA | 0.981 | 0.982 | 0.974 |
| LIME | 0.987 | 0.988 | 0.982 |
| SHAP | 0.961 | 0.980 | 0.947 |
|  | 1000 Iterations | | |
|  | F1 | AUC | Acc |
| MDA | 0.982 | 0.967 | 0.974 |
| LIME | 0.988 | 0.967 | 0.982 |
| SHAP | 0.969 | 0.964 | 0.956 |

(d) Boston Housing Price (Regression)

|  | Default | | |
|---|---|---|---|
|  | MAE | MSE | $R^2$ |
| MDA  | 3.39           | 22.97 | 0.694 |
| LIME | 3.61           | 23.42 | 0.687 |
| SHAP | 3.23           | 22.07 | 0.705 |
|      | 100 Iterations |       |       |
|      | MAE            | MSE   | $R^2$ |
| MDA  | 2.48           | 12.68 | 0.826 |
| LIME | 2.52           | 14.04 | 0.807 |
| SHAP | 2.52           | 13.56 | 0.814 |
|      | 1000 Iterations |      |       |
|      | MAE            | MSE   | $R^2$ |
| MDA  | 2.15           | 8.19  | 0.891 |
| LIME | 2.15           | 8.19  | 0.891 |
| SHAP | 2.05           | 8.14  | 0.891 |

## 6  Application to Trading Strategy Meta-labeling

In this section, we apply feature selection to a data set with the sign of the actual historical returns of a trading strategy as target variable. We want to improve its trading performance by selecting the stable features and use them to predict whether each trade of the strategy will be profitable. If a loss is predicted, we will veto the trading strategy's entry signal. This process has been called meta-labeling by [7].

### 6.1  Description

The dataset contains 464 transactions dated from 2013 to 2019 and 153 features including various market indicators such as implied and realized volatility. The target variable is binary: 1 if the trade is profitable, 0 otherwise. The transactions prior to 2018 form the train set and the ones from Jan 2018 to Oct 2019 form the test set.

### 6.2  Feature Selection

With the number of features from 5 to 152, LIME is shown in Figure 9 to be more stable than MDA and SHAP on the validation set. The 24 features with LIME importance scores greater than their mean will be used in the predictive model.

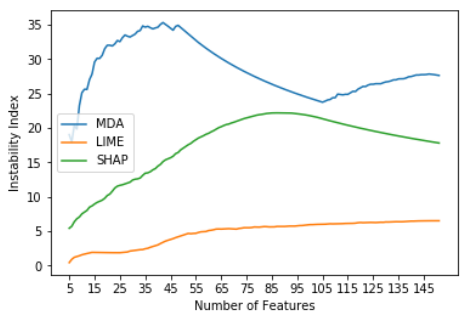

Figure 9: Instability index comparison on Trading dataset

Table 3 summarizes the prediction performance of the various feature selection algorithms. Similar to the results from the other data sets, all feature selection algorithms outperform predictions using all features and the predictive performance differences among the various selection algorithms are insignificant.

Table 3: Summary Table for Prediction Performance (Trading Dataset)

|      | Validation Set |       |       |
|------|----------------|-------|-------|
|      | F1             | AUC   | Acc   |
| MDA  | 0.636          | 0.707 | 0.644 |
| LIME | 0.633          | 0.687 | 0.633 |
| SHAP | 0.667          | 0.700 | 0.667 |
| All  | 0.591          | 0.680 | 0.600 |
|      | Test Set       |       |       |
|      | F1             | AUC   | Acc   |
| MDA  | 0.613          | 0.623 | 0.600 |
| LIME | 0.629          | 0.625 | 0.609 |
| SHAP | 0.625          | 0.630 | 0.628 |
| All  | 0.594          | 0.582 | 0.538 |

### 6.3  Backtest Performance

We compare the backtest trading performance on the test set based on the Sharpe ratio, with and without feature selection, and with the actual historical performance without meta-labeling. We build 100 different random forests, each trained with a different random seed, but all with the same selected features based on the procedure described in Section 6.2. We backtest our trading strategy subject to the meta-label predictions from each random forest and obtain 100 different Sharpe ratios. The histograms of these Sharpe ratios are shown in Figure 10, without LIME feature selection and with. (The red vertical lines in the histograms indicate the mean Sharpe ratios.) Table 4 shows the original strategy without meta-labeling has a Sharpe ratio of 0.36. The mean Sharpe ratio increases to 0.74 when meta-labeling without feature selection is implemented, and it increases to 0.83 when feature selection is implemented.

We can also compare the test set trading performance based on the cumulative returns in Figure 11. The strategy with feature selection has higher average (over different random forests) cumulative return.

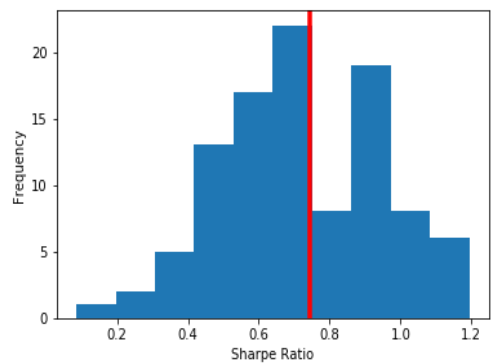

(a)  Strategy without Feature Selection

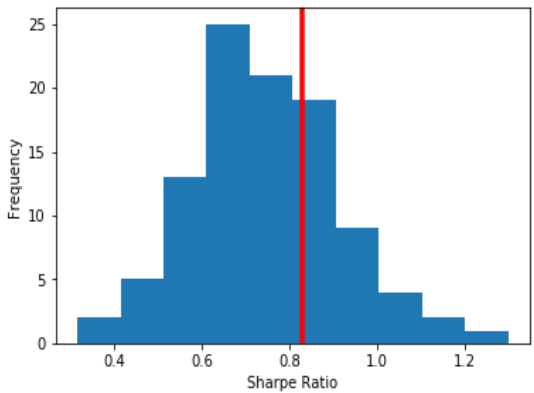

(b)   Strategy with Feature Selection

Figure 10: Sharpe Ratio Comparison

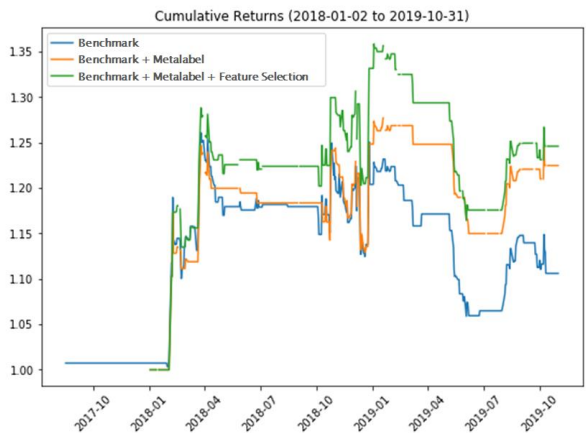

Figure 11: Return Comparisons

Table 4: Sharpe Ratio and Cumulative Returns

|  | Original Strategy | Without Feature Selection | With Feature Selection |
|---|---|---|---|
| Sharpe Ratio | 0.360 | 0.743 | 0.829 |
| Cumulative Return | 0.056 | 0.097 | 0.105 |

## 7  Conclusions

We propose a ranked-based 'instability index' to measure the stability of feature selection algorithms. With this metric, MDA, LIME and SHAP are compared in multiple datasets. We find LIME and SHAP to be more stable than MDA, and LIME is at least as stable as SHAP for the top ranked features. Hence LIME is best suited for human interpretation of a machine learning model. The predictive performance of a model with feature selection improves over a model with no feature selection on synthetic and public datasets, but the three feature selection algorithms' predictive performances do not differ significantly. Furthermore, we show that none of the algorithms converges to a single feature set even if the number of random iterations is large, and this isn't solely due to the substitution effect. However, since the instability index of all the algorithms decreases with increasing number of iterations, we can determine a minimum number of iterations when stability plateaus. On the other hand, we find that high feature stability does not necessarily improve predictive performance. Applying LIME to our trading strategy, both Sharpe ratio and cumulative return of the strategy were improved out-of-sample.

ACKNOWLEDGMENTS

We thank Radu Ciobanu, Tho Do, and Roger Hunter for many useful suggestions and technical assistance.